\documentclass[a4paper]{article}
\usepackage[left=2.5cm, right=2.5cm, top=3cm, bottom=3cm]{geometry}
\usepackage[T1]{fontenc}
\usepackage{polski}
\usepackage[utf8]{inputenc}
\usepackage{amsthm}
\usepackage{amsmath}
\usepackage{amssymb}
\usepackage{graphicx}
\usepackage{array}

\newcommand*{\EN}{\mathbb{N}}
\newcommand*{\ER}{\mathbb{R}}

\newcommand*{\prob}{\mathbb{P}}

\newcommand*{\mieszanka}{\sum_{i = 1}^{K} p_i(\mathbf{x} | \theta_i) P_i}
\newcommand*{\iks}{\mathbf{x}}
\newcommand*{\mi}{\mathbf{m}_i}

\begin{document}

\title{Samoorganizujące się sieci mieszankowe w~reprezentacji fotografii
cyfrowych w~skali szarości}

\author{Patryk Filipiak\\Instytut Informatyki, Uniwersytet Wrocławski\\patryk.filipiak@ii.uni.wroc.pl}

\maketitle

\begin{abstract}
Sieci Kohonena (SOM) są najczęściej wykorzystywanym narzędziem w celu tzw.
uczenia bez nadzoru. Dlatego też doczekały się wielu modyfikacji i adaptacji. 
Niniejsza praca poświęcona jest samoorganizującym się sieciom mieszankowym (SOMN), 
będącym istotnym rozwinięciem pierwotnej idei Kohonena. Zdolność SOMN do efektywnego 
uczenia się dowolnego rozkładu statystycznego ukazana została na przykładzie
fotografii cyfrowych w skali szarości. Dowolny obraz cyfrowy w skali szarości
może być przybliżony za pomocą skończonej mieszanki gaussowskiej, której parametry dobierane
są automatycznie w procesie uczenia SOMN. W niniejszej publikacji przedstawiona
została grupa przykładów takiego wykorzystania SOMN przy użyciu zaimplementowanej w tym celu aplikacji.
\end{abstract}

\section{Wstęp}

Reprezetacja w skali szarości obrazu o rozdzielczości $M \times N$ pikseli
sprowadza się do przypisania każdemu z $M \cdot N$ punktów wartości natężenia jego jasności. Wielkość tę zwyczajowo poddajemy dyskretyzacji
do wartości całkowitych z przedziału od 0 do 255, co umożliwia przechowanie jej w
dokładnie jednym bajcie pamięci komputera.

Określamy funkcję jasności $l : \{0, \ldots, M-1\} \times \{0, \ldots, N-1\}
\longrightarrow \{0, \ldots, 255\}$, która każdemu pikselowi obrazu $(x, y) \in
\{0, \ldots, M-1\} \times \{0, \ldots, N-1\}$ przyporządkowuje dyskretną wartość
natężenia jego jasności.

Niech:
\[ L = \sum_{x = 0}^{M-1} \sum_{y = 0}^{N-1} l(x,y) . \]

Wówczas funkcja $l' = \frac{l}{L}$ jest dyskretyzacją funkcji gęstości pewnego
rozkładu wektora losowego w~przestrzeni dwuwymiarowej. Możemy zatem postrzegać
obraz w kryteriach rozkładu statystycznego.

Niech $\iks$ będzie wektorem losowym w przestrzeni $d$-wymiarowej 
$\Omega \subseteq \ER^d$ ($d \geqslant 1$). Mówimy, że \textbf{wektor losowy
$\iks$ ma rozkład w postaci skończonej mieszanki} (ang. \emph{finite mixture distribution}), jeżeli funkcja gęstości
jego rozkładu jest następująca:
\begin{equation}
\label{equation:finiteMixtureDistribution}
p(\iks) = p_1(\iks) P_1 + \ldots + p_K(\iks) P_K  \qquad
(\iks \in \Omega, \; K \geqslant 1),
\end{equation}
gdzie
\[ P_i \geqslant 0, \; i = 1, \ldots, K; \quad \sum_{i = 1}^{K} P_i = 1 \]
oraz
\[ p_i(\cdot) \geqslant 0, \; i = 1, \ldots, K; \quad \int_{\Omega}
p_i(\iks) \, d\iks = 1. \]

Zmienne $P_1, \ldots, P_K$ nazywać będziemy \textbf{wagami}, zaś funkcje
$p_1(\cdot), \ldots, p_K(\cdot)$ --- \textbf{składnikami mieszanki}.
Dla mieszanek jednorodnych (tzn. takich, których składniki są funkcjami
gęstości tego samego typu) wygodnie będzie zapisać Równanie
\ref{equation:finiteMixtureDistribution} w postaci
\begin{equation}
\label{equation:homogenousMixtureDistribution}
p(\iks | \Theta) = \mieszanka,
\end{equation}
gdzie $\theta_i$ są wektorami parametrów $i$-tego rozkładu, zaś $\Theta =
(\theta_1, \ldots, \theta_K)$ .

W niniejszej pracy rozważać będziemy \emph{mieszanki gaussowskie} zadane wzorem 

\begin{equation}
\forall_{i = 1, \ldots, K} \quad p_i(\iks | \theta_i) = \frac{1}{(2
\pi)^\frac{d}{2} \cdot \left| \Sigma_i \right| ^ \frac{1}{2}} \cdot \exp \left[
{-\frac{1}{2} (\iks-\mathbf{m_i})^T \Sigma_{i}^{-1} (\iks-\mathbf{m_i})} \right],
\end{equation}
gdzie $\theta_i = \{ \theta_{i1}, \theta_{i2} \} = \{ \mathbf{m_i}, \Sigma_i \}$
to (kolejno) wektor wartości średnich oraz macierz kowariancji rozkładu
normalnego.

\section{Klasteryzacja za pomocą sieci Kohonena (SOM)}

Niech $d \geqslant 1$ oraz $\{ \iks \, ; \; \iks = (x_1,
\ldots, x_d) \in \Omega \subseteq \ER^d \}$ będzie próbką z pewnego
rozkładu prawdopodobieństwa (ciągłego lub dyskretnego), zaś
$\{ \mathbf{m} \} = \{ \mi \in \ER^d \, ; \; i = 1, \ldots, K \}$
ustalonym zbiorem tzw. \emph{wektorów kotwicowych}.

\textbf{Teselacją Voronoi'a} \cite{coolen} przestrzeni $\Omega \subseteq \ER^d$
nazywamy procedurę jej podziału na $K$ wypukłych podzbiorów $V_i(\{ \mathbf{m} \})$
wyznaczonych przez wektory $\mi \in \ER^d$ następująco:
\begin{equation}
\forall_{i = 1, \ldots, K} \qquad V_i(\{ \mathbf{m} \}) = \{ \iks \in
\Omega \, ; \; \forall_{j \neq i} \; \| \iks - \mi \| < \|
\iks - \mathbf{m}_j \| \} ,
\end{equation}
gdzie $\| \cdot \|$ oznacza normę euklidesową. Uzyskany podział nazywamy
\emph{mozaiką Voronoi'a}.

Zauważmy, jakie znaczenie odgrywa odpowiedni dobór wektorów kotwicowych. Aby
wynikowa mozaika Voronoi'a była reprezentatywna dla wyjściowego rozkładu
$p(\iks)$ i użyteczna dla potrzeb klasteryzacji, musimy zadbać o to, by gęstość
rozmieszczenia wektorów $\mi$ była proporcjonalna do gęstości $p(\iks)$ w danym regionie. Innymi słowy ---
chcemy, aby: 
\[ \forall_{i \neq j} \qquad \prob(\iks \in V_i(\{ \mathbf{m} \})) = \prob(\iks \in V_j(\{ \mathbf{m} \})) . \]

\begin{figure}[!h]
\begin{center}
\begin{picture}(250,200)(0,0)
\put(15,15){\includegraphics[width=8cm]{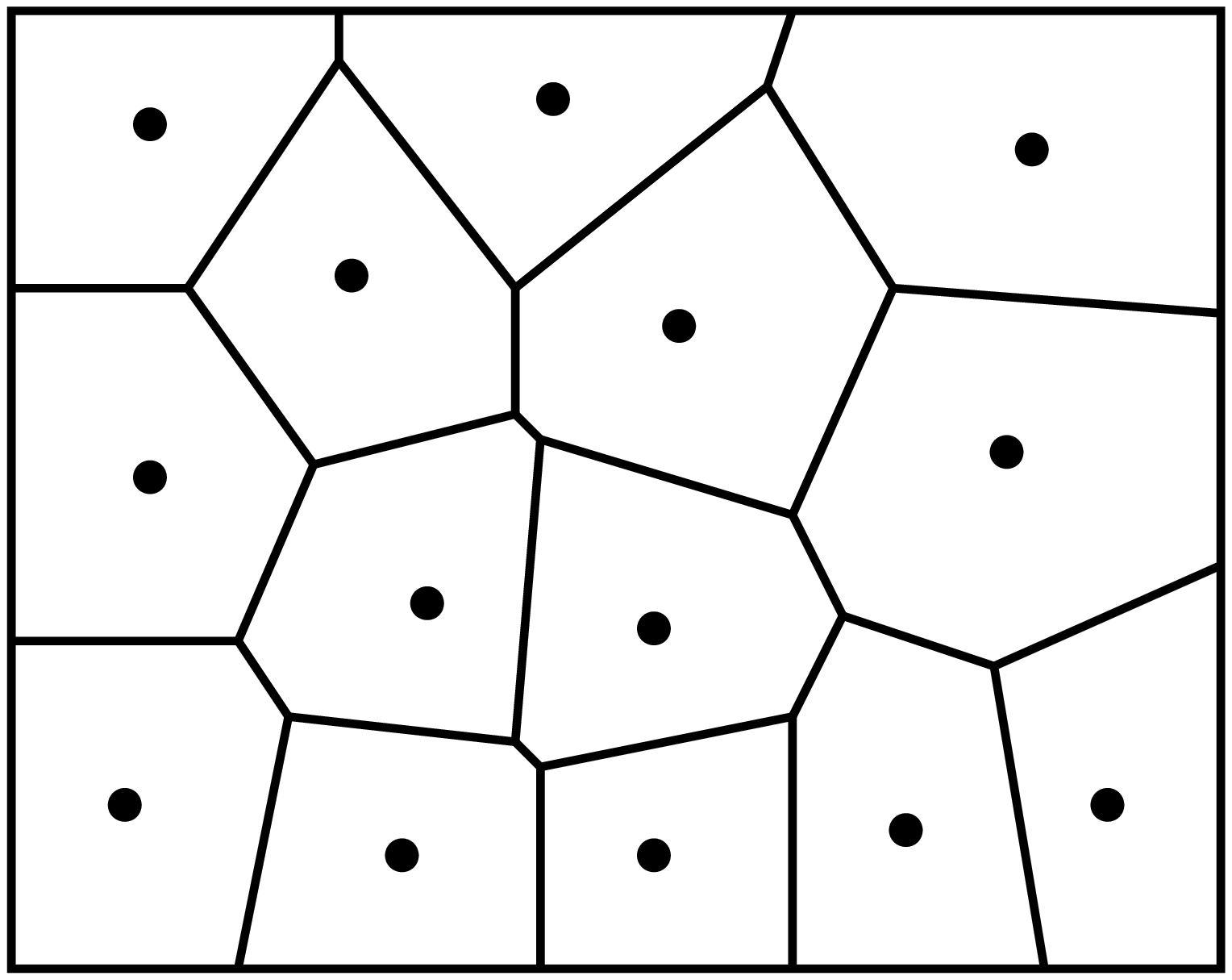}}
\put(130,5){\makebox(0,0){$x_1$}}
\put(5,105){\makebox(0,0){$x_2$}}
\put(213, 118){\makebox(0,0){$\mi$}}
\end{picture}
\end{center}
\caption{Przykładowa mozaika Voronoi'a dla $d = 2$. Rysunek
zaczerpnięto z \cite{coolen}.}
\end{figure} 

Sieci Kohonena (SOM, od ang. Self-organizing Maps \cite{som,
kohonen}) realizują koncepcję tzw. \emph{samoorganizacji topologicznej}.
Nacisk kładziony jest bowiem nie tylko na rozmieszczenie wektorów
kotwicowych $\mi \in \ER^d$ traktowanych niezależnie, ale 
również ich wzajemne położenie.

\vspace{10pt}
Niech $G$ będzie dowolnym grafem spójnym, którego wierzchołkami są
wszystkie $K$ wektory kotwicowe $\mi$. Niech ponadto $d(i, j)$ będzie
pewną grafową miarą odległości w $G$. Określmy funkcję $\widetilde{h} :
\ER_{+}\cup\{0\} \longrightarrow [0, 1]$ o własnościach:
\begin{enumerate}
  \item[(a)] $\widetilde{h}$ jest ściśle malejąca,
  \item[(b)] $\widetilde{h}(0) = 1$,
  \item[(c)] $\lim_{z \rightarrow \infty} \widetilde{h}(z) = 0$.
\end{enumerate}
Dla tak zadanego odwzorowania $\widetilde{h}$ zdefiniujmy \emph{funkcję
sąsiedztwa} (ang. \emph{neighbourhood function}) $h : \{ (i, j) \, ; \; i, j = 1, \ldots, K,
\; i \neq j \} \longrightarrow \ER$ następująco:
\begin{equation}
\label{equation:neighbourhoodFunction}
\forall_{i \neq j} \quad h(i, j) = \widetilde{h} \left( \frac{d(i, j)}{\sigma}
\right) ,
\end{equation}
gdzie $\sigma > 0$ jest pewnym parametrem algorytmu. W każdej iteracji wektor
$\iks$ leżący w $i$-tej komórce Voronoi'a porusza wszystkie
$j$-te wektory kotwicowe, dla których $d(i, j) < \sigma$.

\subsection{Algorytm SOM}
\begin{itemize}
  \item \emph{Dane:} Liczba $K \in \EN$, skończony zbiór wektorów
  losowych $\iks \in \Omega \subseteq \ER^d$ zgodny z rozkładem prawdopodobieństwa $p(\iks)$, funkcja sąsiedztwa $h(i, j)$ oraz parametry $\sigma, \eta$.
  \item \emph{Wynik:} Zbiór wektorów kotwicowych $\{ \mathbf{m} \} = \{ \mi \in \ER^d \, ; \; i = 1, \ldots, K \}$ tworzących
  strukturę grafu, reprezentatywny dla $p(\iks)$.
\end{itemize}
Wybierz losowo wszystkie wektory $\mi \in \ER^d$, a następnie
powtarzaj:
\begin{enumerate}
  \item Wylosuj wektor $\iks \in \Omega$ zgodnie z rozkładem
  $p(\iks)$.
  \item Znajdź komórkę Voronoi'a zawierającą $\iks$, to znaczy wyznacz
  indeks $i$ taki, że:
\[ \forall_{j \neq i} \quad \| \iks - \mi \| < \|
\iks - \mathbf{m}_j \| . \]
  \item Przesuń wszystkie wektory kotwicowe $\mathbf{m}_j$ w stronę
  $\iks$ według następującej formuły:
\begin{equation}
\label{equation:somFormula}
\mathbf{m}_j \leftarrow \mathbf{m}_j + \eta (\iks - \mathbf{m}_j) h(i,
j) .
\end{equation}
\end{enumerate}

\section{Samoorganizujące się sieci mieszankowe}

\emph{Samoorganizujące się sieci mieszankowe}
(ang. \emph{Self-Organizing Mixture Networks}, w skrócie \emph{SOMN})
\cite{somn} są uogólnieniem pojęcia \emph{Bayesian Self-Organizing Maps}
(\emph{BSOM}) \cite{bsom}. Cechuje je dwuwartstwowa struktura, pozwalająca połączyć zalety sieci Kohonena (SOM) z możliwością uczenia macierzy kowariancji oraz wag dla każdej ze składowych mieszanki.

\subsection{Architektura sieci}

\begin{figure}[!h]
\begin{center}
\begin{picture}(250,230)(0,0)
\put(15,5){\includegraphics[width=8cm]{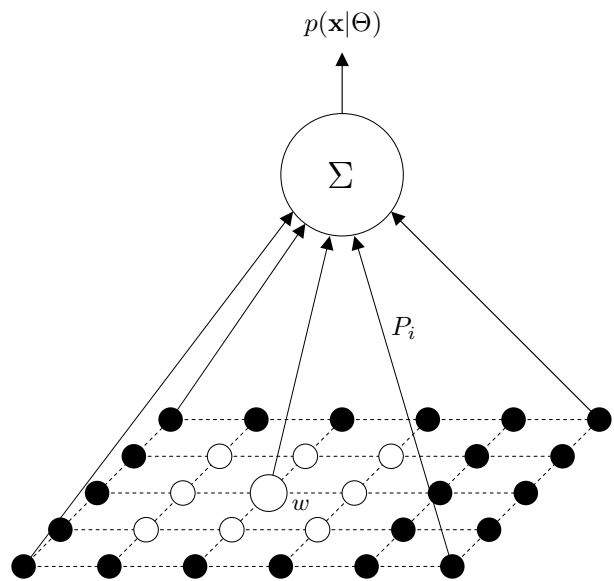}}
\put(141, 215){\makebox(0,0){$p(\iks | \Theta)$}}
\put(140, 158){\makebox(0,0){\Large{$\Sigma$}}}
\put(164, 100){\makebox(0,0){$P_i$}}
\put(125, 33){\makebox(0,0){\small{$w$}}}
\end{picture}
\end{center}
\caption{Schematyczne przedstawienie SOMN
zaczerpnięte z \cite{somn}.}
\label{figure:layers}
\end{figure} 

Na Rysunku~\ref{figure:layers} przedstawiono schematycznie
dwuwarstwową strukturę sieci. Niższa warstwa funkcjonuje na zasadach bardzo zbliżonych do
sieci Kohonena. Wyższa natomiast powstaje poprzez
sumowanie wszystkich węzłów z uwzględnieniem wag $P_i$, z jakimi węzły
te występują w wynikowej mieszance (por. Równanie
\ref{equation:homogenousMixtureDistribution}), dając w wyniku wartość
$p(\iks | \Theta)$ dla dowolnego $\iks \in \Omega$.

\begin{figure}[h!]
\begin{center}
\begin{tabular}{ccc}
\includegraphics[width=4cm]{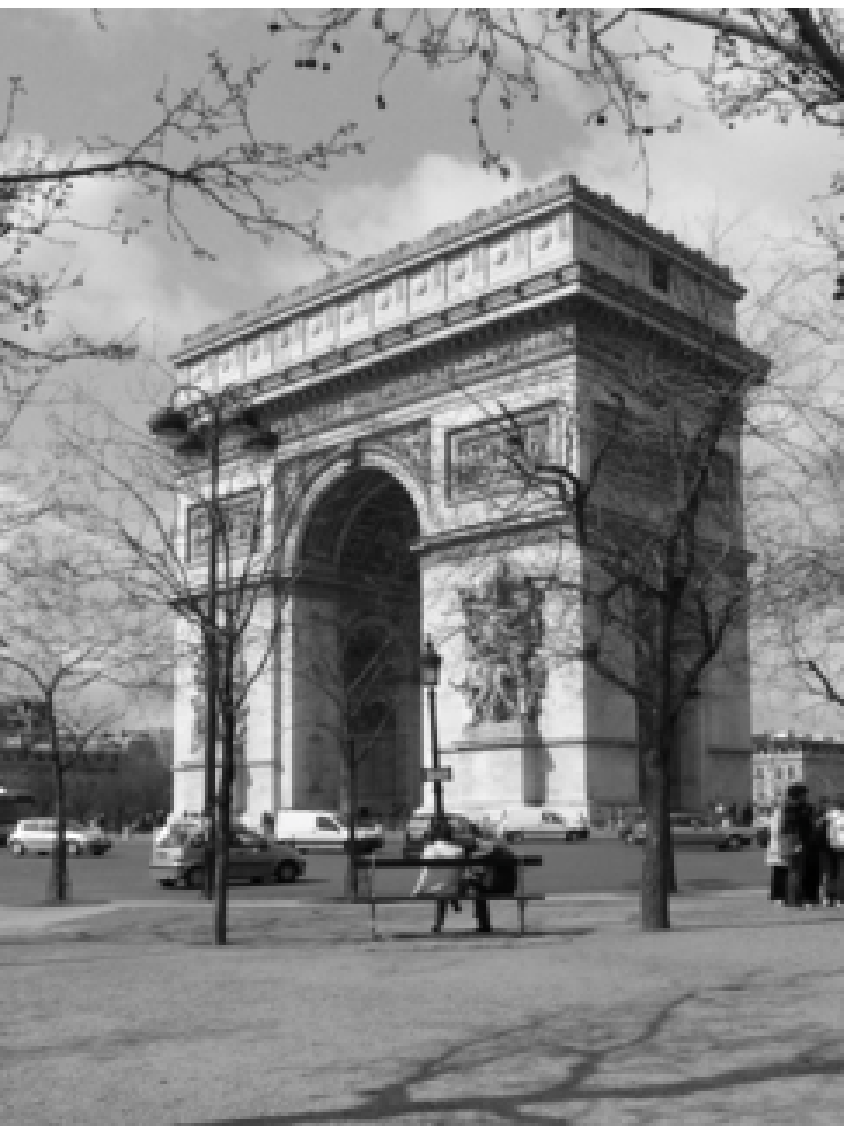}
& 
\includegraphics[width=4cm]{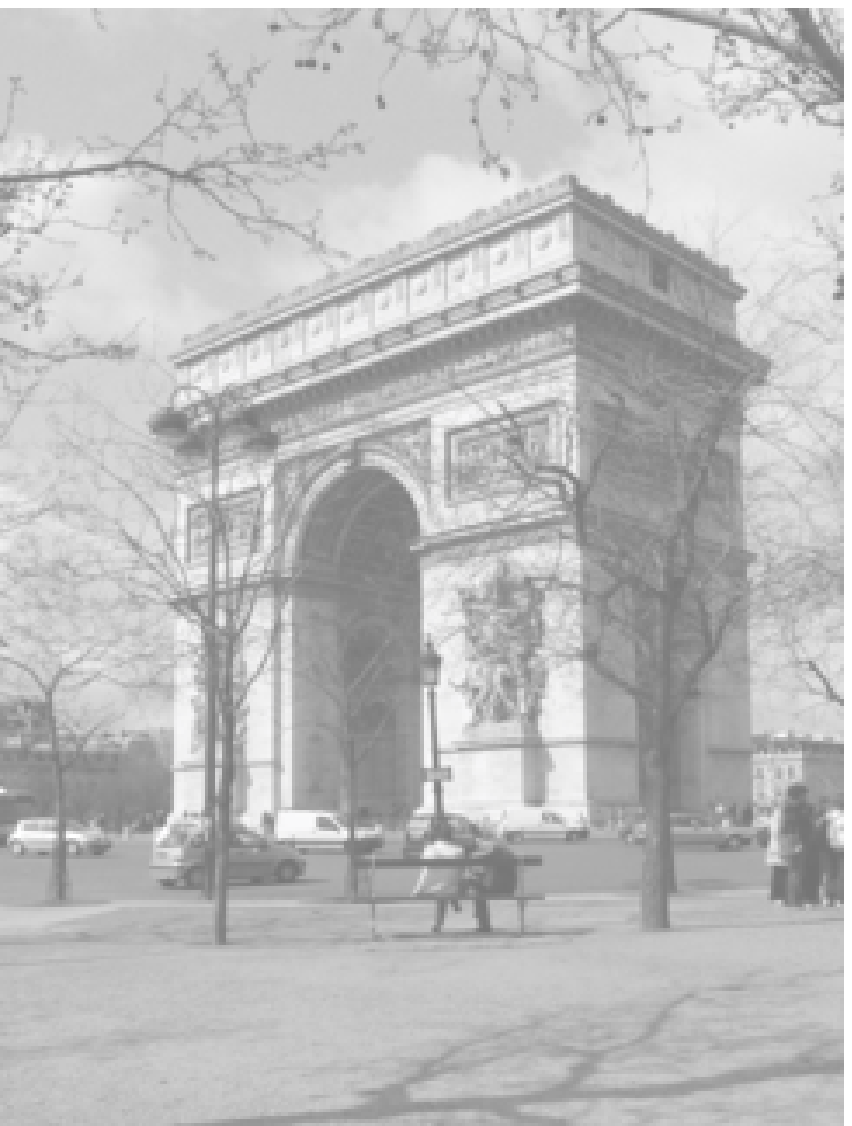}
& 
\includegraphics[width=4cm]{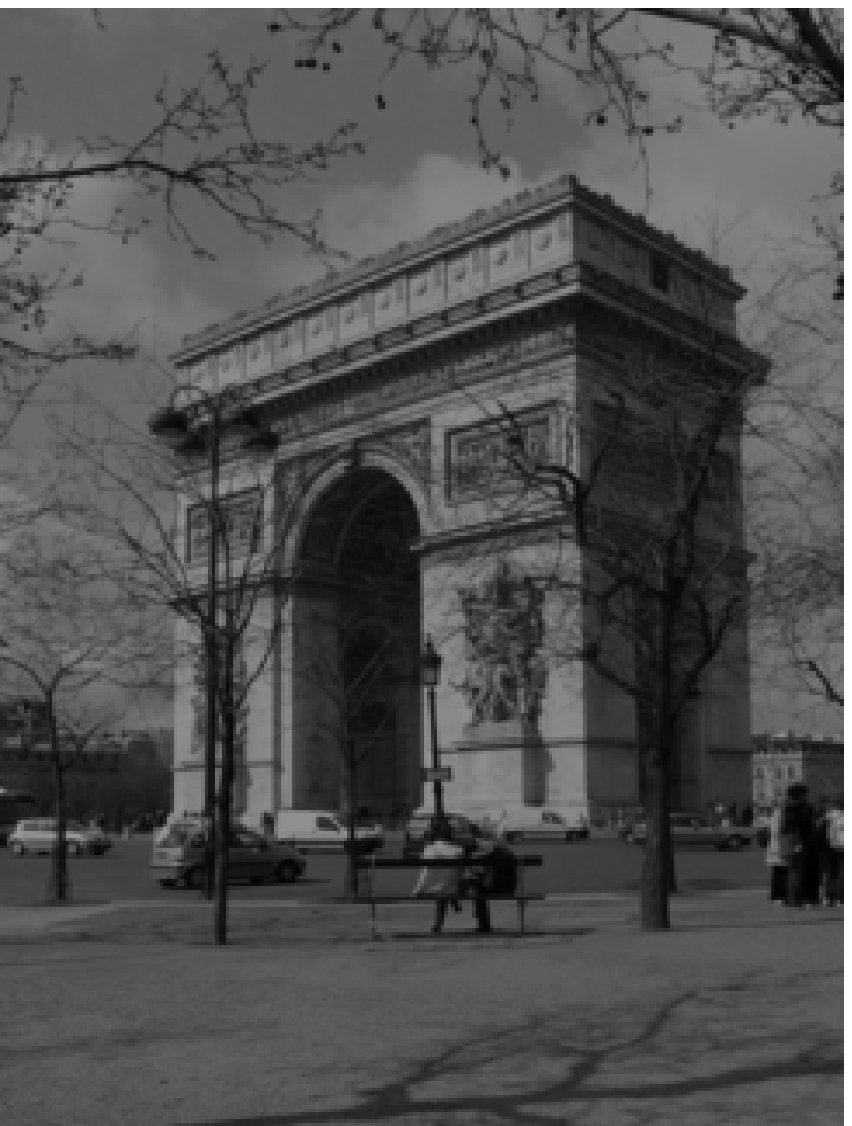}
\\
(a) & (b) & (c) \\
\end{tabular}
\caption{Odwzorowanie jasności obrazu: (a) obraz wyjściowy, (b) i (c)
przykłady niepoprawnego odwzorowania.}
\label{figure:brigthnessFit}
\end{center}
\end{figure}

W praktycznym zastosowaniu SOMN do reprezentacji obrazów cyfrowych,
przekształcenie funkcji gęstości rozkładu na wartość jasności każdego kolejnego
piksela jest niewystarczające.
Ignoruje ono bowiem informację o średniej jasności całego obrazu, traktując
jednakowo obrazy utrzymane w jasnych jak i ciemnych odcieniach, odwzorowując
jedynie kontrast pomiędzy najjaśniejszym i najciemniejszym pikselem. Prowadzić
to może do błędnej reprezentacji wyjściowego obrazu (por. Rysunek
\ref{figure:brigthnessFit}).

Celem uniknięcia powyższego błędu, należy dla wejściowego obrazu o
rozdzielczości $M \times N$ wyznaczyć sumę jasności wszystkich $M \cdot N$
pikseli. Pamiętając o wspomnianej wcześniej kwestii rozbieżności w reprezentacji odcieni, tzn. $l(\cdot, \cdot) = 0$
oznacza odcień najciemniejszy, zaś gęstość $p(\cdot | \cdot) = 0$ --- odcień najjaśniejszy ---
obliczymy w istocie sumę różnic postaci:
\[ L' = \sum_{x = 0}^{M-1} \sum_{y = 0}^{N-1} (255 - l(x,y)) . \]

\subsection{Uczenie SOMN}

Uczenie SOMN jest procesem iteracyjnym. W chwili inicjalizacji sieć składa się
z $\widetilde{K}$ węzłów, gdzie $\widetilde{K} \geqslant K$. Jeżeli wartość $K$ jest znana \emph{a priori},
przyjmujemy zwykle $\widetilde{K} = K$. W przeciwnym razie szacujemy ją z góry,
aby uniknąć błędu wynikającego ze zbyt niewielkiej liczby składników mieszanki,
co w oczywisty sposób obniżałoby dokładność przybliżenia rozkładu.
 
W każdym z $T \in \EN$ kroków uczenia sieci
losujemy próbkę $\iks(t)$ zgodnie z przybliżanym rozkładem
$p(\iks)$, w konsekwencji czego nieznacznie modyfikujemy jej stan. Aktualną
postać mieszanki oznaczmy jako:
\begin{equation}
\hat{p}(\iks | \hat{\Theta}) = \sum_{i = 1}^{\widetilde{K}}
\hat{p}_i(\iks | \hat{\theta}_i) \hat{P}_i , 
\end{equation}
gdzie $\hat{\Theta} = (\hat{\theta}_1, \ldots, \hat{\theta}_{\widetilde{K}})$.

Niech $t = 1, \ldots, T$ będzie numerem bieżącej iteracji. Wektorowi $\iks(t)$
przyporządkowujemy węzeł, dla którego wartość prawdopodobieństwa
warunkowego (\emph{a posteriori}) obliczanego według formuły:
\begin{equation}
\hat{P}(i | \iks) = \frac{\hat{P}_i \hat{p}_i (\iks |
\hat{\theta}_i)}{\hat{p}(\iks | \hat{\Theta})}
\end{equation}
jest największa. Węzeł taki zwyczajowo nazywamy \emph{zwycięzcą} (ang.
\emph{winner}). Na Rysunku \ref{figure:layers} przedstawiony jest jako
największe niewypełnione koło oznaczone literą $w$. Pozostałe niewypełnione
koła symbolizują węzły znajdujące się w bezpośrednim sąsiedztwie zwycięzcy.

\subsection{Algorytm SOMN}

\begin{itemize}
  \item \emph{Dane:} Liczba $\widetilde{K} \in \EN$,
  skończony zbiór wektorów losowych $\iks \in \Omega \subseteq \ER^d$ zgodny z rozkładem
  prawdopodobieństwa $p(\iks)$, odległość grafowa $d(i, j)$, ciąg
  nierosnący $(a_t)_{t = 1}^{\infty}$ zbieżny do zera oraz skończony ciąg
  niemalejący $(\delta_t)_{t = 1}^{T}$ o tej własności,
  że $\delta_T = 0$.
  \item \emph{Wynik:} Graf węzłów sieci (z
  parametrami $\hat{P}_i, \hat{\theta_i}$,
  gdzie $\hat{\theta}_i = \{ \hat{\mathbf{m}}_i, \hat{\Sigma}_i \}$ dla $i =
  1, \ldots, \widetilde{K}$) reprezentatywny dla $p(\iks)$.
\end{itemize}
Zainicjuj dowolną metodą wszystkie parametry $\hat{P}_i, \hat{\theta}_i$, a
następnie dla $t = 1, \ldots, T$ powtarzaj:
\begin{enumerate}
  \item Wylosuj wektor $\iks \in \Omega$ zgodnie z rozkładem
  $p(\iks)$.
  \item Znajdź indeks $i \in \{ 1, \ldots, \widetilde{K} \}$ węzła, dla którego
  wartość: \[ \hat{P}(i | \iks) = \frac{\hat{P}_i \hat{p}_i (\iks |
\hat{\theta}_i)}{\hat{p}(\iks | \hat{\Theta})} \]
  jest największa.
  \item Dla wszystkich węzłów $j \in \{ 1, \ldots, \widetilde{K} \}$ takich, że
  $d(i, j) \leqslant \delta(t)$ przypisz:
\begin{eqnarray*}
\hat{\mathbf{m}}_j &\leftarrow& \hat{\mathbf{m}}_j + a(t) \hat{P}(j | \iks)
\left[ \iks - \hat{\mathbf{m}}_j \right] , \\
\hat{\Sigma}_j &\leftarrow& \hat{\Sigma}_j + a(t) \hat{P}(j | \iks) \left\{
\left[ \iks - \hat{\mathbf{m}}_j \right] \left[ \iks -
\hat{\mathbf{m}}_j \right]^T - \hat{\Sigma}_j \right\} , \\
\hat{P}_j &\leftarrow& \hat{P}_j + a(t) \left[ \hat{P}(j | \iks) -
\hat{P}_j \right] .
\end{eqnarray*}
  \item Dla wszystkich węzłów $j = 1, \ldots, \widetilde{K}$ wykonaj
  normalizację:
  \[ \hat{P}_j(t) \: \leftarrow \: \frac{\hat{P}_j(t)}{\sum_{k =
  1}^{\widetilde{K}} \hat{P}_k(t)} . \]
\end{enumerate}

\section{Przykłady}

Testy zostały wykonane przy wykorzystaniu programu (stanowiącego integralną
cześć pracy \cite{mgr}) implementującego opisany wariant algorytmu SOMN. Program
ten dostarcza możliwość skonfigurowania następujących parametrów uczenia sieci:
Suwak \emph{Learn} pozwala wybrać wartość z przedziału $0,01 \div 1,00$ z dokładnością $0,01$. Wyrazy ciągu współczynników sterujących tempem uczenia dla wektorów średnich $\hat{\mathbf{m}}_i$ oraz macierzy kowariancji $\hat{\Sigma}_i$ określone są wzorem: $a(t) = Cooling\,factor (t) \cdot
Learn$. Dla zniwelowania efektu zbyt gwałtownych wahań wag węzłów SOMN w
procesie uczenia, wprowadzono dodatkowy parametr \emph{Weight} z przedziału
$0,00001 \div 0,00100$ określany z dokładnością $0,00001$. Ciąg współczynników
determinujących tempo uczenia wag węzłów wyrażony jest więc wzorem: $\alpha(t) =
a(t) \cdot Weight$.


W pierwszej kolejności posłużymy się fotografią gmachu węgierskiego parlamentu w
Budapeszcie:

\begin{figure}[h!]
\begin{center}
\includegraphics[width=6cm]{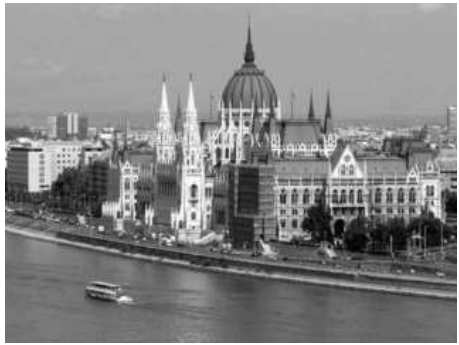}
\caption{Obraz wejściowy w skali szarości ($640 \times 480$ pikseli).}
\label{figure:budapestSource}
\end{center}
\end{figure}

\begin{figure}[h!]
\begin{center}
\begin{tabular}{ccc}
\includegraphics[width=6cm]{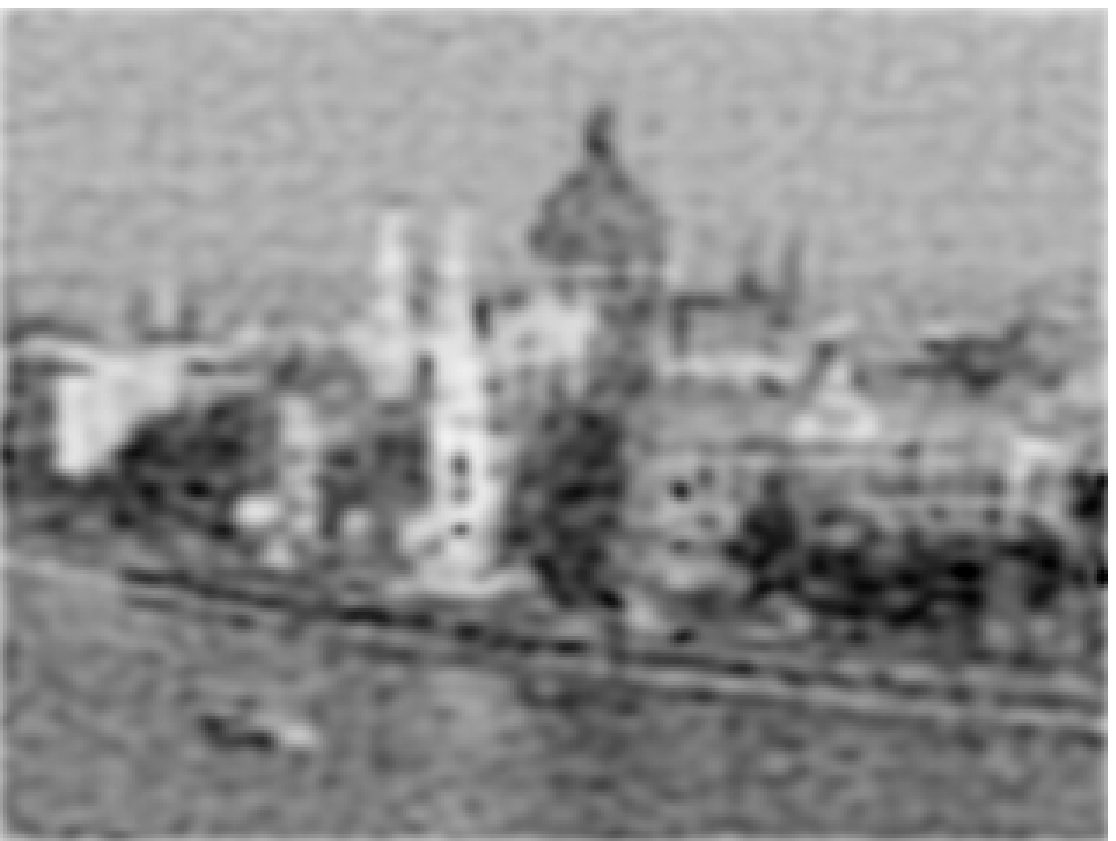}
& &
\includegraphics[width=6cm]{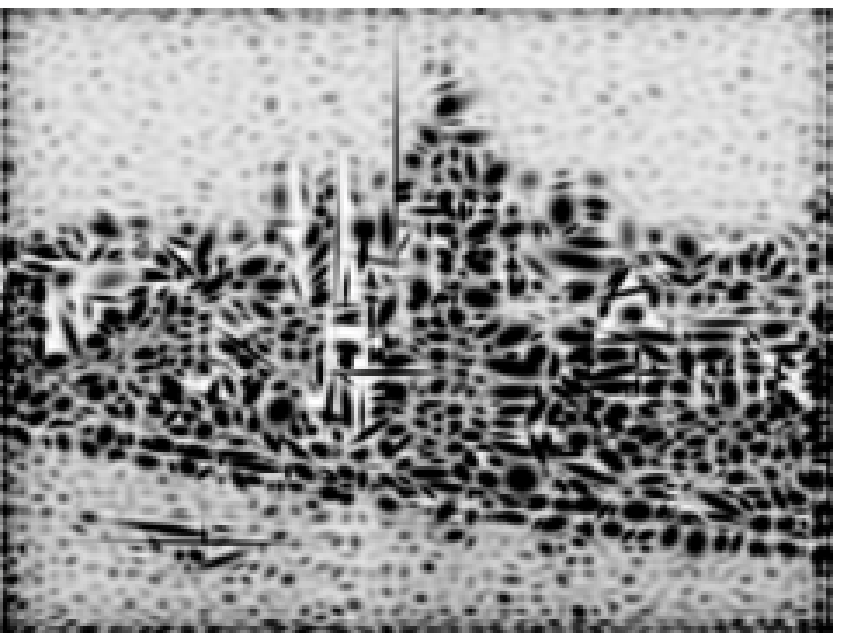}
\\
(a) & & (b) \\
\end{tabular}
\caption{Sieć rozmiaru $100 \times 100$ węzłów: (a) po 100 tys. iteracji,
(b) po 5 mln. iteracji ze zbyt wysokimi parametrami uczenia. }
\label{figure:budapestBadParameters}
\end{center}
\end{figure}

Zdjęcie przesycone jest detalami architektonicznymi, dlatego jego poprawne
odwzorowanie wymaga precyzyjnego dobrania parametrów. Rysunek
\ref{figure:budapestBadParameters}(a) pokazuje, że 100 tysięcy iteracji to
zdecydowanie zbyt mało, by oddać szczegóły budowli. Obraz jest mocno rozmyty i
nieczytelny. Z kolei Rysunek \ref{figure:budapestBadParameters}(b) przedstawia
wynik działania SOMN po pięciu milionach iteracji z parametrami: $Learn = 0,15
$ oraz $Weight = 0,00005$, które w przypadku tego obrazu okazały się być zbyt
wysokie. Zważywszy, że uzyskanie wyniku w niniejszym przypadku pochłonęło nieco
ponad 7 godzin, widzimy, że właściwe dobranie parametrów staje się procesem
czasochłonnym.

\vspace{10pt}

\newpage
Tabela \ref{table:budapestSummary} przedstawia zestawienie najlepszych
uzyskanych wyników. Rezultat dla sieci $200 \times 200$ wygląda
satysfakcjonująco, jednak należy zwrócić uwagę, że 5 milionów iteracji algorytmu
pochłonęło znacznie ponad jedną dobę.

\vspace{30pt}
\begin{table}[h!]
\begin{center}
\begin{tabular}{|r|c|c|}
\hline
& \multicolumn{2}{c|}{Rozmiar sieci} \\
\cline{2-3}
\raisebox{2ex}{Liczba iteracji} & $100 \times 100$ & $200 \times 200$ \\
\hline
&&\\[-10pt]
\raisebox{7ex}{1 000 000}
&
\includegraphics[width=6cm]{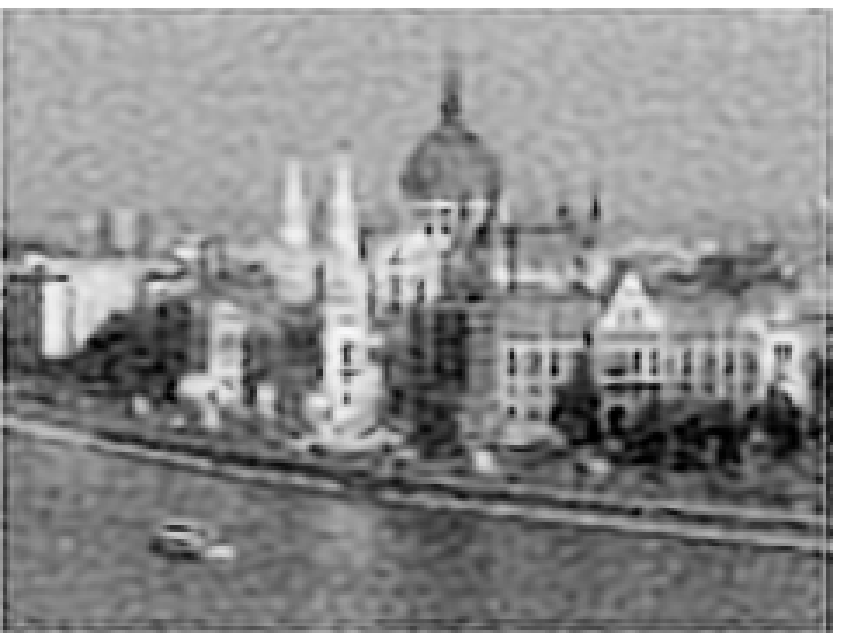}
&
\includegraphics[width=6cm]{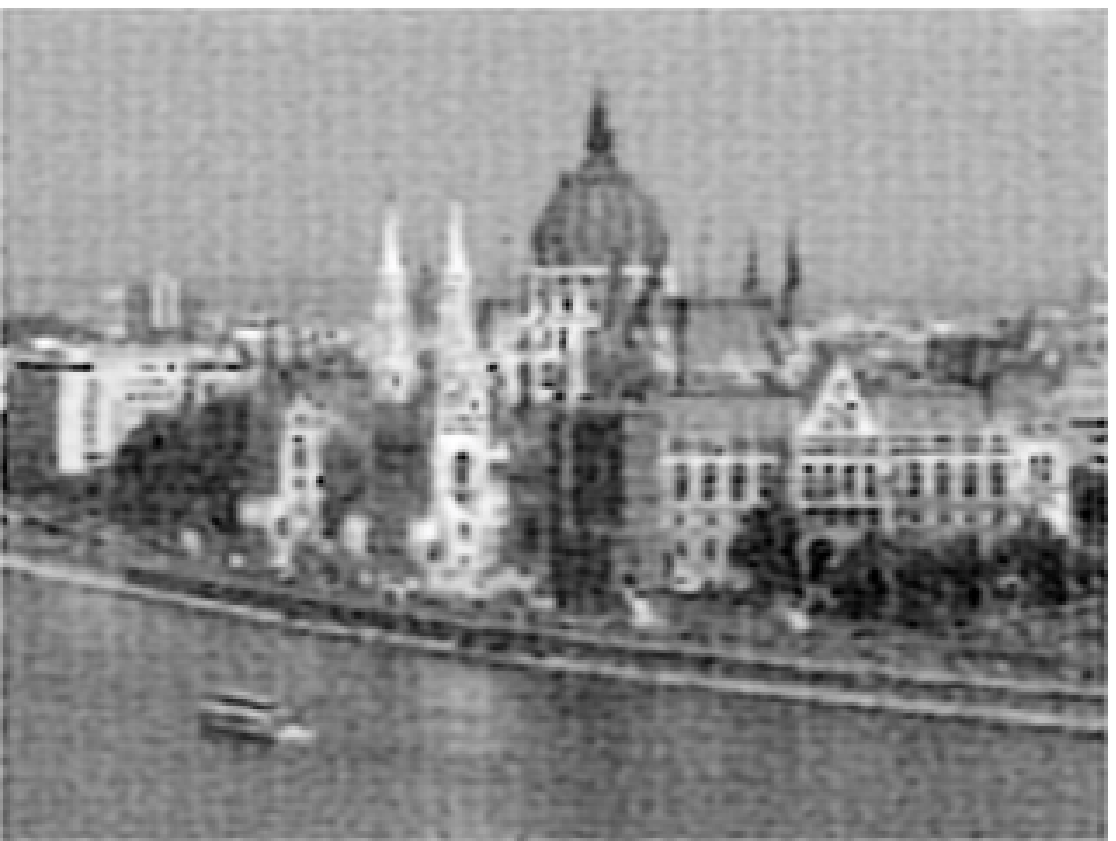}
\\
& 1 godz. 22 min. & 6 godz. 29 min. \\ 
\hline
&&\\[-10pt]
\raisebox{7ex}{5 000 000}
&
\includegraphics[width=6cm]{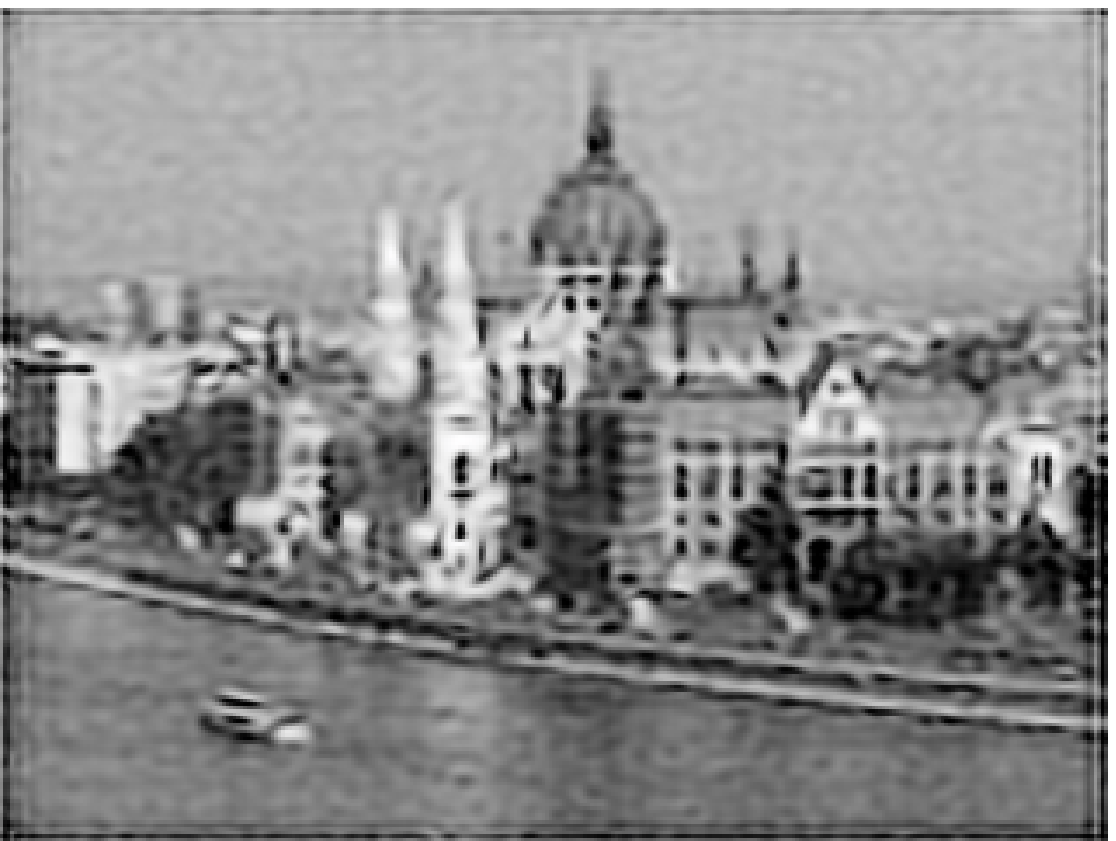}
&
\includegraphics[width=6cm]{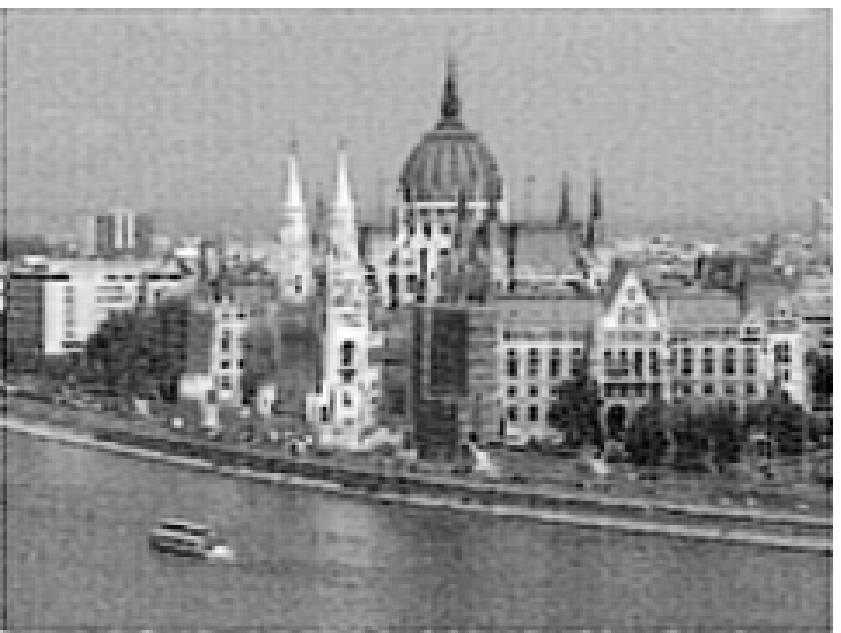}
\\
& 7 godz. 16 min. & 31 godz. 21 min. \\ 
\hline
\end{tabular}
\end{center}
\caption{Zestawienie rezultatów i czasów wykonywania algorytmu SOMN dla
podanych rozmiarów sieci oraz liczby iteracji na przykładzie fotografii
gmachu węgierskiego parlamentu.}
\label{table:budapestSummary}
\end{table}

\begin{figure}[h!]
\begin{center}
\includegraphics[width=5cm]{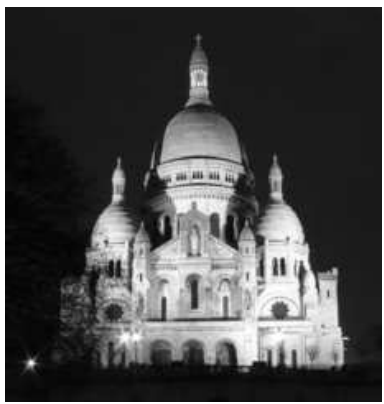}
\caption{Obraz wejściowy w skali szarości ($460 \times 480$ pikseli).}
\label{figure:parisSource}
\end{center}
\end{figure}

\newpage
Zupełnie inny charakter cechuje kolejną fotografię przedstawiającą paryską
bazylikę Sacré-Cœur (Rysunek \ref{figure:parisSource}). Zdjęcie zostało wykonane nocą,
zatem odwzorowanie ciemnego nieba skontrastowanego z jasną elewacją kościoła stanowi kompletnie odmienne
zadanie dla SOMN w stosunku do poprzedniego przykładu.

\vspace{10pt}

Trudność w odwzorowaniu dużych pól jednakowego odcienia skutkuje występowaniem
artefaktów, widocznych na poniższym rysunku: 

\vspace{15pt}
\begin{figure}[h!]
\begin{center}
\begin{picture}(320,170)(0,0)
\put(0,30){\includegraphics[width=5cm]{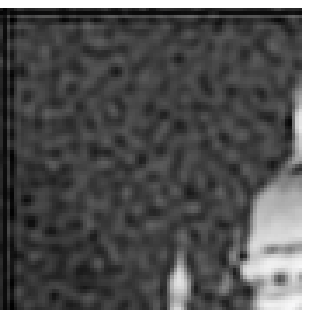}}
\put(200,0){\includegraphics[width=4cm]{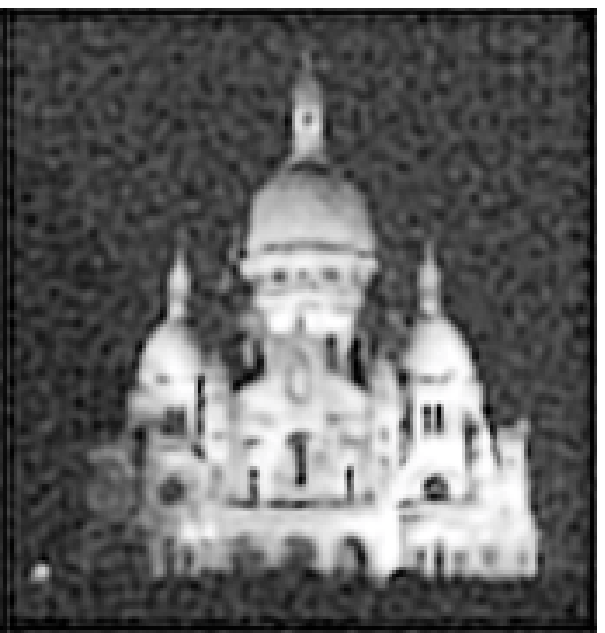}}
\put(145,150){\line(1,0){83}}
\put(228,150){\line(0,-1){25}}
\put(195,59){\dashbox{2}(65,65){}}
\end{picture}
\caption{Artefakty widoczne na dużych polach tego samego odcienia.}
\label{figure:artifacts}
\end{center}
\end{figure}

Artefakty szczególnie widoczne są w przypadku sieci o rozmiarach $100 \times
100$ węzłów, mniej zaś dla rozmiaru $200 \times 200$ (por. Tabela
\ref{table:parisSummary}).

\newpage
\begin{table}[t!]
\begin{center}
\begin{tabular}{|r|c|c|}
\hline
& \multicolumn{2}{c|}{Rozmiar sieci} \\
\cline{2-3}
\raisebox{2ex}{Liczba iteracji} & $100 \times 100$ & $200 \times 200$ \\
\hline
&&\\[-10pt]
\raisebox{7ex}{1 000 000}
&
\includegraphics[width=5cm]{images_paris_100-100_1mln_0-15_0-00005.eps}
&
\includegraphics[width=5cm]{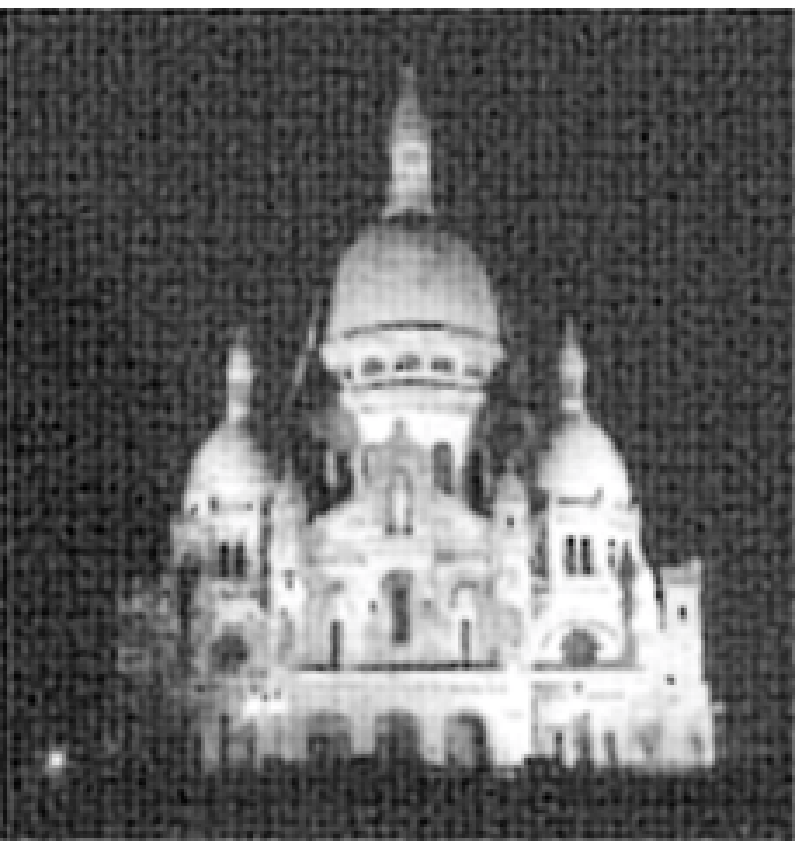}
\\
& 1 godz. 4 min. & 5 godz. 57 min. \\ 
\hline
&&\\[-10pt]
\raisebox{7ex}{5 000 000}
&
\includegraphics[width=5cm]{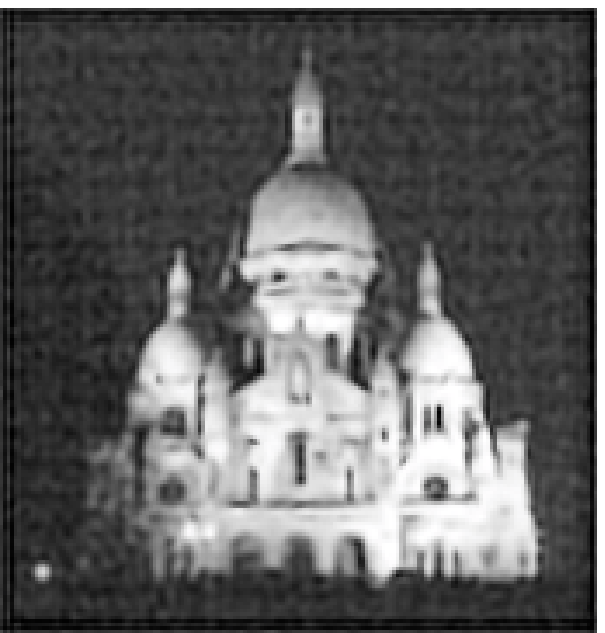}
&
\includegraphics[width=5cm]{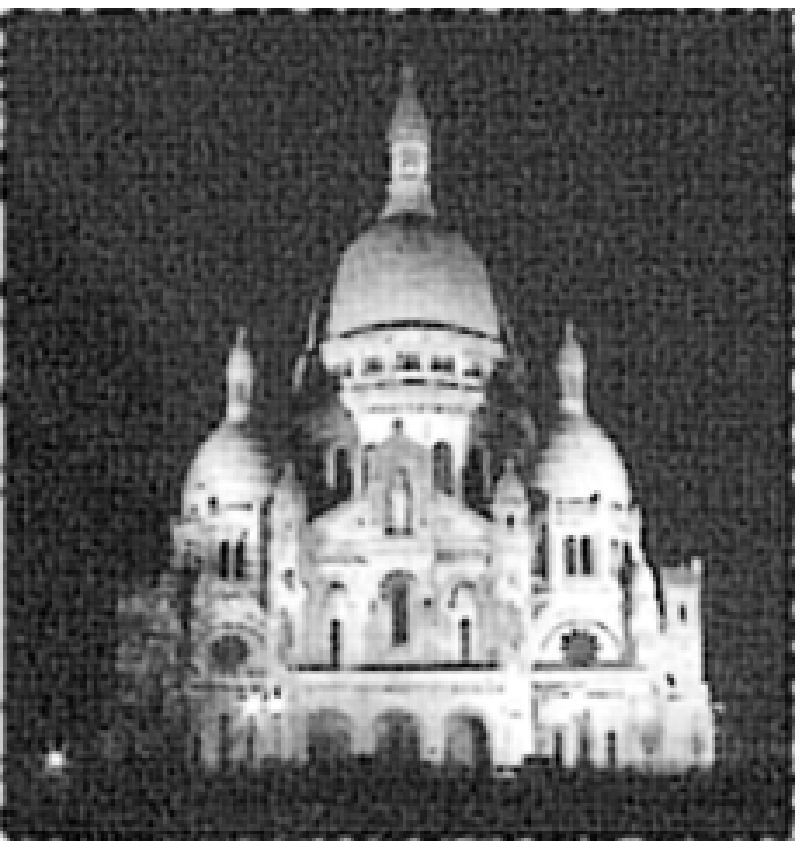}
\\
& 6 godz. 32 min. & 28 godz. 11 min. \\ 
\hline
\end{tabular}
\end{center}
\caption{Zestawienie rezultatów i czasów wykonywania algorytmu SOMN dla
podanych rozmiarów sieci oraz liczby iteracji na przykładzie fotografii
bazyliki Sacré-Cœur.}
\label{table:parisSummary}
\end{table}

\begin{figure}[h!]
\begin{center}
\includegraphics[width=6cm]{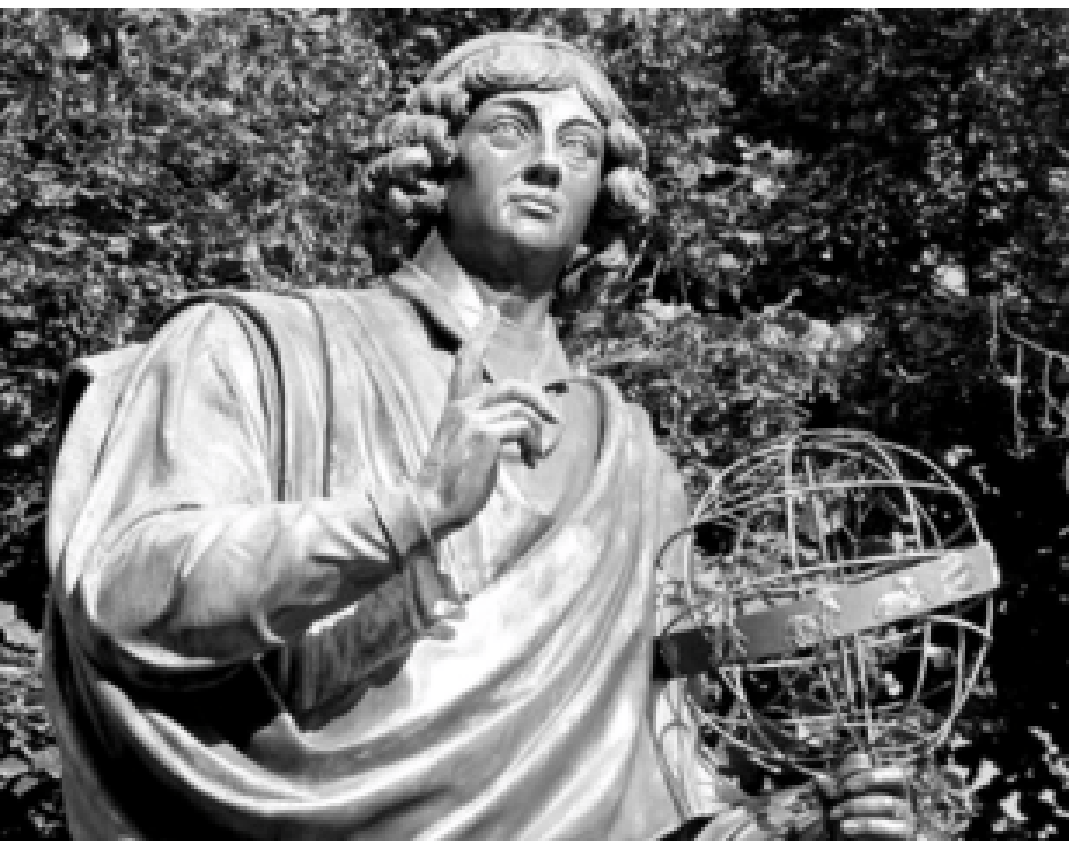}
\caption{Obraz wejściowy w skali szarości ($640 \times 480$ pikseli).}
\label{figure:kopernikSource}
\end{center}
\end{figure}

Jako ostatni przykład weźmiemy pod uwagę fotografię pomnika Mikołaja Kopernika w
Toruniu (Rysunek \ref{figure:kopernikSource}). Podobnie jak w poprzednich
przypadkach, najlepsze odwzorowanie detali uzyskujemy po pięciu milionach
iteracji algorytmu dla sieci o rozmiarze $200 \times 200$ węzłów. Odwzorowanie
detali rysów twarzy, jak również liści drzew na drugim planie jest bliskie
oryginałowi. Brak rozległych pól jednakowego odcienia powoduje, że na uzyskanym
obrazie nie odnajdujemy niepożądanych artefaktów. Zestawienie uzyskanych
rezultatów przedstawiono w Tabeli \ref{table:kopernikSummary}.

\vspace{10pt}

\begin{table}[t!]
\begin{center}
\begin{tabular}{|r|c|c|}
\hline
& \multicolumn{2}{c|}{Rozmiar sieci} \\
\cline{2-3}
\raisebox{2ex}{Liczba iteracji} & $100 \times 100$ & $200 \times 200$ \\
\hline
&&\\[-10pt]
\raisebox{7ex}{1 000 000}
&
\includegraphics[width=6cm]{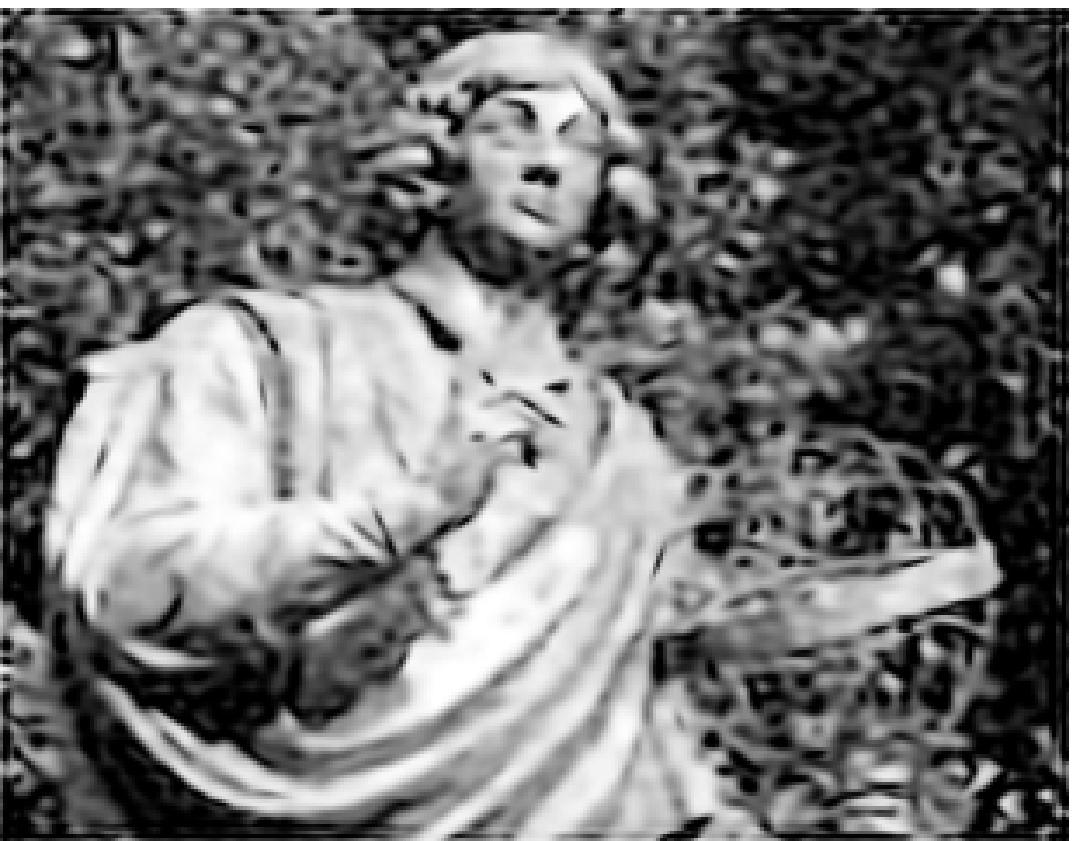}
&
\includegraphics[width=6cm]{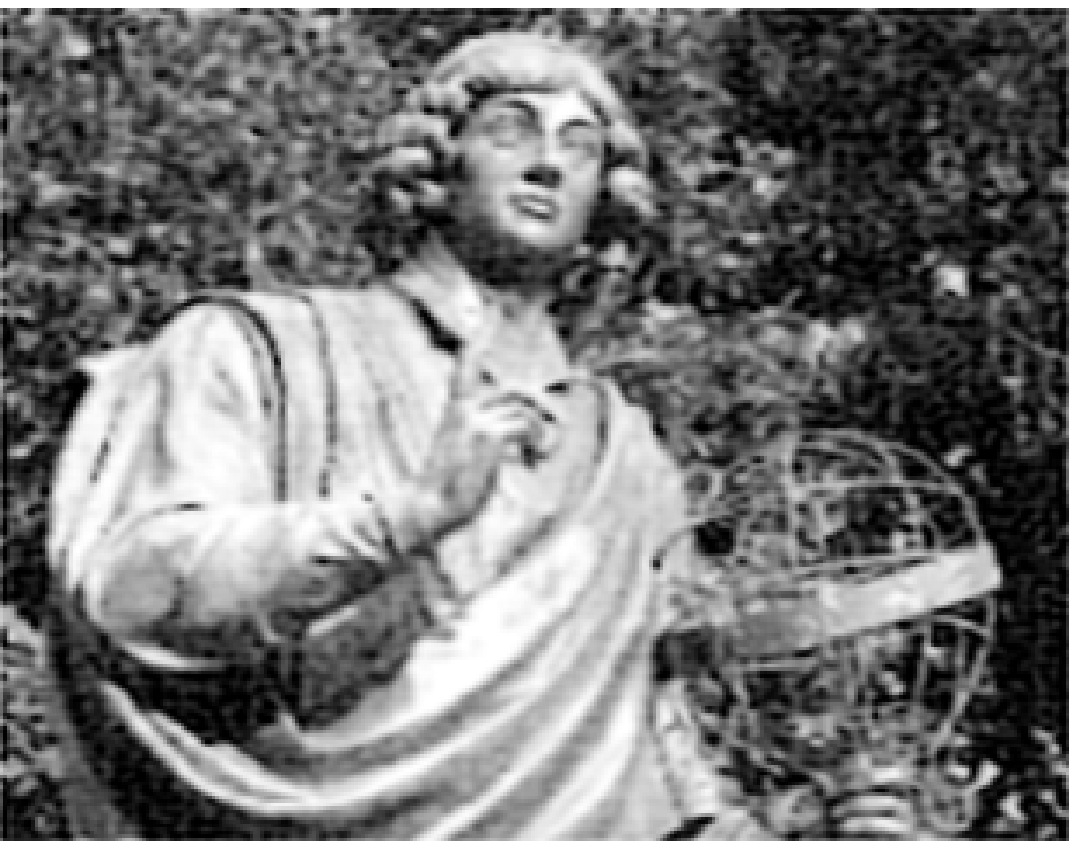}
\\
& 1 godz. 31 min. & 6 godz. 27 min. \\ 
\hline
&&\\[-10pt]
\raisebox{7ex}{5 000 000}
&
\includegraphics[width=6cm]{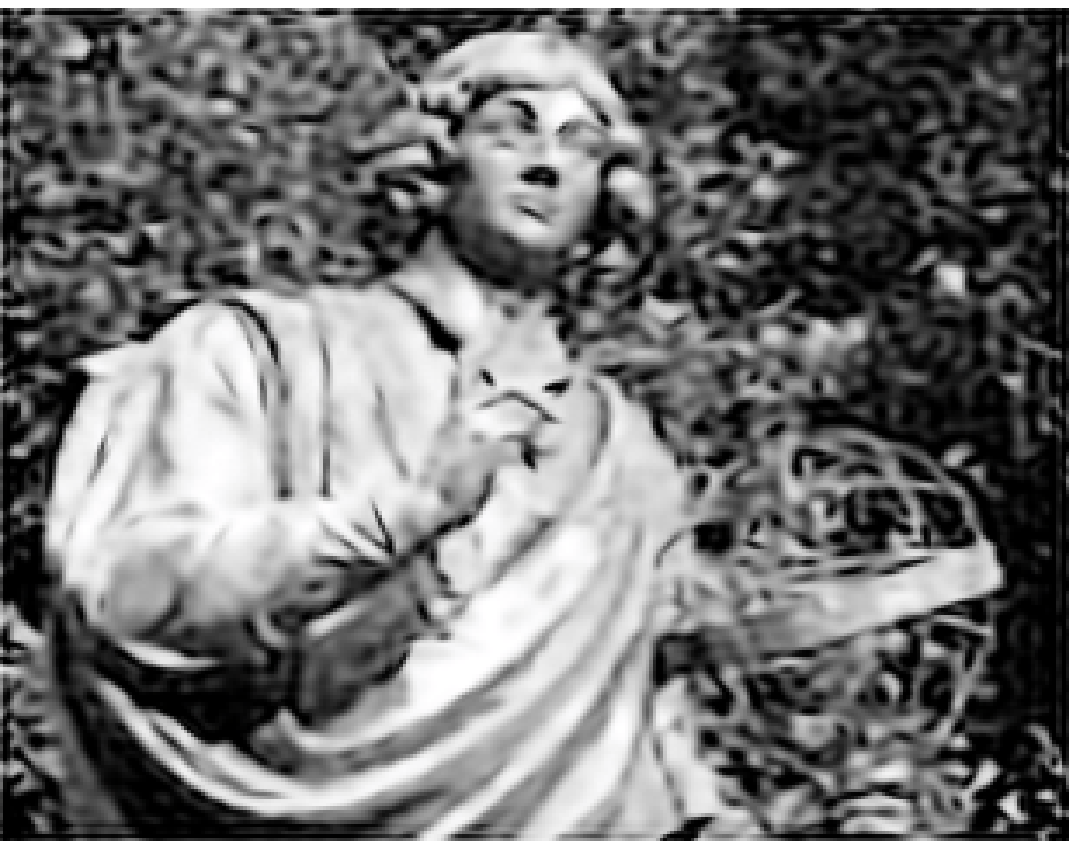}
&
\includegraphics[width=6cm]{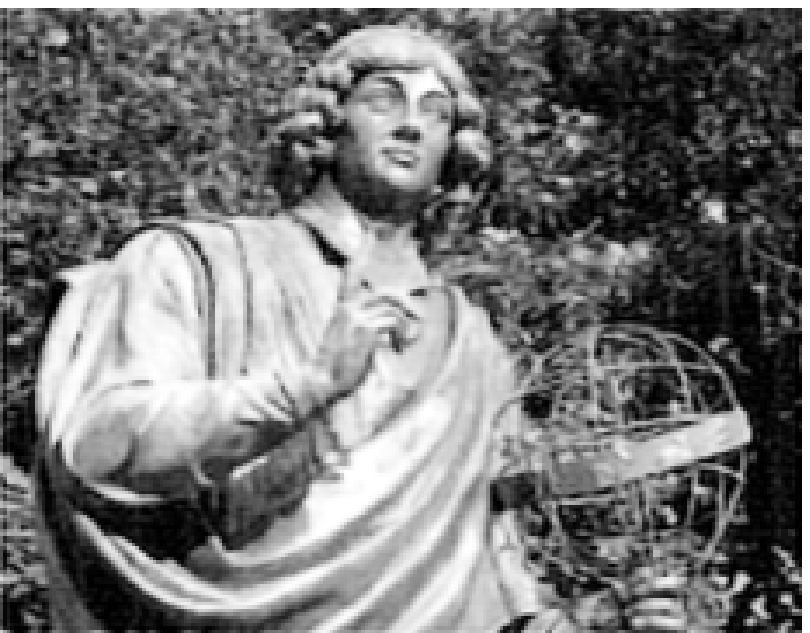}
\\
& 7 godz. 8 min. & 31 godz. 55 min. \\ 
\hline
\end{tabular}
\end{center}
\caption{Zestawienie rezultatów i czasów wykonywania algorytmu SOMN dla
podanych rozmiarów sieci oraz liczby iteracji na przykładzie fotografii
pomnika Mikołaja Kopernika.}
\label{table:kopernikSummary}
\end{table}

\section{Konkluzje}

Niniejsza publikacja oparta jest w znacznej mierze na pracy magisterskiej autora
\cite{mgr} napisanej pod kierunkiem prof. Tomasza Schreibera na Uniwersytecie
Mikołaja Kopernika w Toruniu.

Jako dane wejściowe dla procesu klasteryzacji z użyciem SOMN wykorzystane
zostały obrazy cyfrowe w skali szarości. Warto zwrócić uwagę, że reprezentacja
obrazu przy pomocy sieci neuronów jest bardzo użyteczna dla potrzeb analizowania wzorców (ang. \emph{pattern matching}), eksploracji
danych (ang. \emph{data mining}) czy kompresji. Przedstawione przykłady
pokazują, że odpowiednio nauczone SOMN mogą być z powodzeniem wykorzystane do
reprezentacji obrazów cyfrowych w skali szarości. 

\bibliographystyle{plain}
\bibliography{bibliografia}

\end{document}